\documentclass[letterpaper, 10pt, conference]{ieeeconf}
\IEEEoverridecommandlockouts
\usepackage{cite}
\usepackage{amsmath,amssymb,amsfonts}
\usepackage{algorithmic}
\usepackage{graphicx}
\usepackage{textcomp}
\usepackage{xcolor}
\usepackage{caption}
\usepackage{subcaption}
\usepackage{hyperref}
\usepackage{cleveref}
\usepackage{multirow}

\usepackage{siunitx}

\usepackage[normalem]{ulem} 
\usepackage[export]{adjustbox}  
\usepackage{array}  
\usepackage[bottom]{footmisc} 

\allowdisplaybreaks

\newcommand*\diff{\mathop{}\!\mathrm{d}}

\title{\LARGE \bf
Robust Task and Motion Planning for Long-Horizon Architectural Construction Planning
}

\author{Valentin N. Hartmann$^{1}$, Ozgur S. Oguz$^{1,2}$, Danny Driess$^{1,2}$, Marc Toussaint$^{1,2}$ and Achim Menges$^{3}$%
\thanks{The research has been supported by the Deutsche Forschungsgemeinschaft (DFG, German Research Foundation) under Germany's Excellence Strategy -- EXC 2120/1 -- 390831618, and the DAAD.}%
\thanks{$^{1}$Machine Learning \& Robotics Lab, University of Stuttgart, Germany
        {\tt\footnotesize \{firstname\}.\{lastname\}@ipvs.uni-stuttgart.de}}%
\thanks{$^{2}$Max Planck Institute for Intelligent Systems, Germany
        }%
\thanks{$^{3}$Institute for Computational Design and Construction, University of Stuttgart, Germany
        {\tt\footnotesize achim.menges@icd.uni-stuttgart.de}}%
}

\begin{document}

\maketitle
\thispagestyle{empty}
\pagestyle{empty}

\begin{abstract}
Integrating robotic systems in architectural and construction processes is of core interest to increase the efficiency of the building industry.
Automated planning for such systems enables design analysis tools and facilitates faster design iteration cycles for designers and engineers. 
However, generic task-and-motion planning (TAMP) for long-horizon construction processes is beyond the capabilities of current approaches.
In this paper, we develop a multi-agent TAMP framework for long horizon problems such as constructing a full-scale building.
To this end we extend the Logic-Geometric Programming framework by sampling-based motion planning, a limited horizon approach, and a task-specific structural stability optimization that allow an effective decomposition of the task.
We show that our framework is capable of constructing a large pavilion built from several hundred geometrically unique building elements from start to end autonomously.
\end{abstract}

\section{INTRODUCTION}

Building construction is the largest industry worldwide. It consumes around 40\% of global resources and energy, and produces 50\% of all waste \cite{unreport17}. 
Yet, current construction processes, their planning and supervision, as well as the building designs are far from optimal.
Building information modeling (BIM), an increasingly employed method in construction, comprises digital tools designed to manage established, primarily manual construction processes with conventional building elements.
We believe that AI and robotics have the potential to revolutionize the area, moving towards integrated reasoning about the robotized construction process jointly with the building design, resource consumption, and uncertainties.

This paper aims to make a first step in this direction.
We address the problem of computing a possible robotic construction sequence for a given building design.
We reason on the kinematic, geometric and static stability level, neglecting the dynamic constraints and actual control problem of execution.
While assuming the final design as given in this work, the long term goal is to reason jointly over the construction process and the building design:
Such a complementary design-for-robotic-assembly framework should suggest modifications that lead to more efficient construction, and help discover and explore construction processes and related possible designs that are beyond current designer's intuition and conventions. 
We aim to think of the design of the architecture and the construction process as a joint problem. 
In its current form, our framework supports an architect in reviewing a potential robotic construction process for his design, possibly yielding insights to modify the design.

\begin{figure}
\centering
\begin{subfigure}[t]{.22\textwidth}
  \centering
  \adjincludegraphics[width=0.9\textwidth, trim={{.2\width} {.07\width} {.2\width} {.05\width}}, clip]{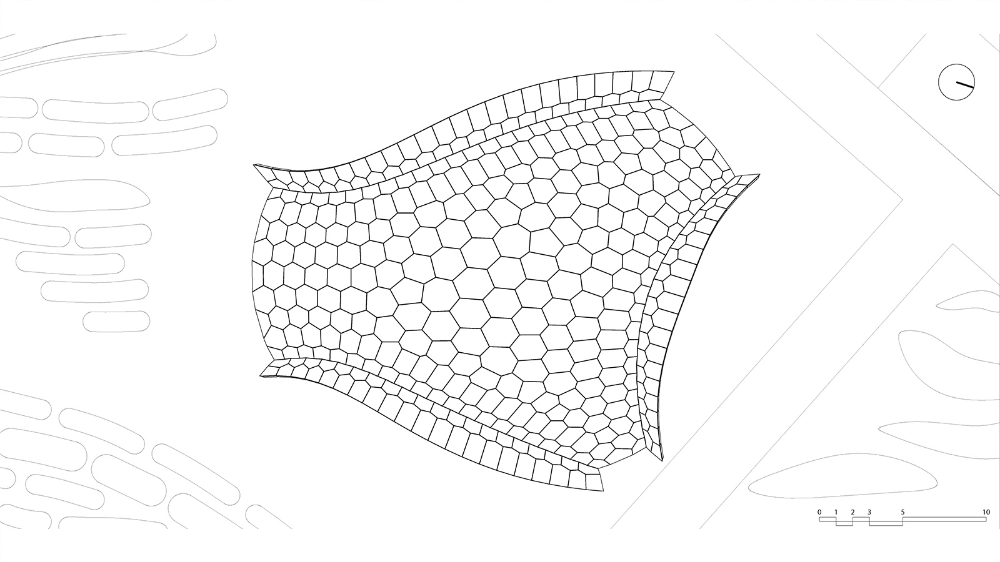}
  \label{fig:Buga_Top}  
  \caption{Schematic top view.}
\end{subfigure}
\hspace{.5em}
\begin{subfigure}[t]{.24\textwidth}
   \centering
  \includegraphics[width=0.9\textwidth]{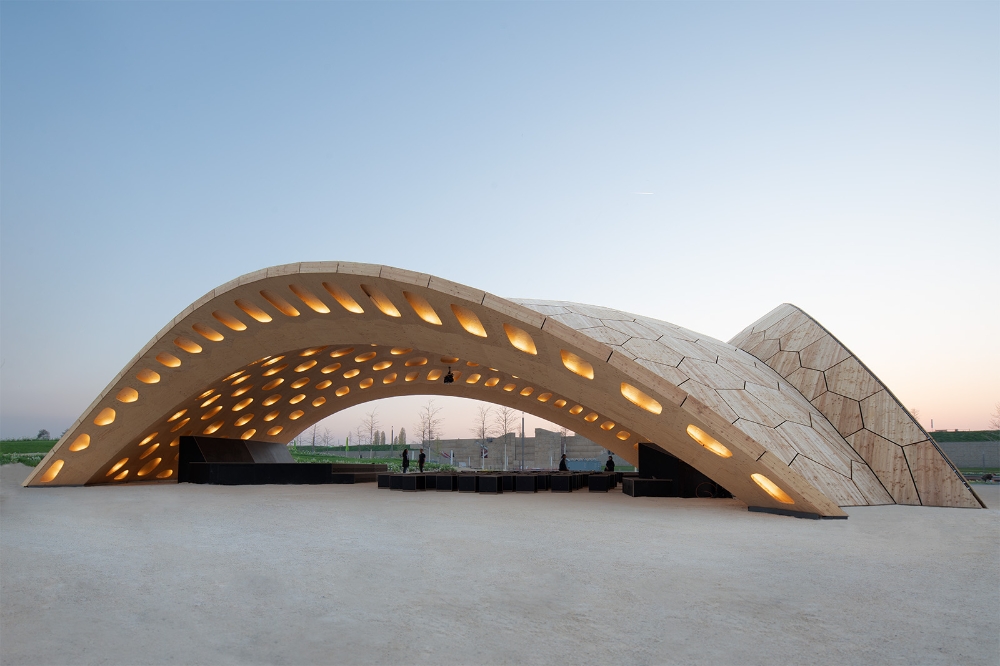}
  \label{fig:Buga_photograph}  
  \caption{Assembled pavilion.}
\end{subfigure}
\caption{BUGA Wood pavilion built by our framework in simulation.}
\label{fig:buga}
\end{figure}

For the demonstration of the planning framework in this paper we consider a long-span pavilion design: the BUGA Wood pavilion (\cref{fig:buga}), which was conventionally built in Heilbronn (DE) at the Bundesgartenschau 2019~\cite{buga2019}.
This example is particularly suitable, as it showcases how the differentiation of building elements, enabled by robotic prefabrication, results in a high-performance, materially efficient structure.
However, the previous assembly and construction remained entirely manual and conventional. 

The problem setting is challenging from the perspective of robot task and motion planning due to several reasons.
The overall construction requires to assemble 376 geometrically unique pieces -- in our model this is done by coordinating 2 robots: one crane and one mobile robot for final placement. 
A core issue when scaling TAMP to long horizons is the trade-off between decomposing the problem and aiming for joint optimality: Previous work on optimization-based TAMP treated the whole manipulation path jointly which will not scale to realistic construction domains.
Additionally, for such long horizon problems, the TAMP-solver needs to be robust over a wide range of arising motion planning problems, while still providing solutions in a reasonable timeframe.
Clearly, we can treat many aspects roughly independently or incrementally provided that certain objectives, e.g., static feasibility, are satisfied.
As such, it should be possible to achieve close to linear scaling of the planning-complexity with the number of required manipulation steps.
We therefore believe that one of the core technical challenges is to identify the (inter)dependence between manipulations/individual pieces, which in turn would determine planning jointly or sequentially.

We propose to follow a limited horizon approach along with task specific objectives as a form of decomposing the problem, and introduce sampling-based approaches in combination with optimization-based methods to form a robust TAMP-solver.
We demonstrate this approach in the Logic-Geometric Programming (LGP) framework introduced in \cite{toussaint2017multi}, which formulates the TAMP as an optimization problem.

The key contributions of this paper are:
\begin{itemize}
    \item A limited (receding) horizon planning approach as a form of decomposition in task and motion planning,
    \item Planning for multiple robots with mixed capabilities,
    \item Introducing methods to robustly solve an LGP problem, such as sampling-based motion planning and stochastic restarts,
    \item Integrating static stability estimates in TAMP planning,
    \item Demonstrating TAMP methods on architectural construction.
\end{itemize}

\section{RELATED WORK}

\subsection{Task and motion planning}

A full discussion of the TAMP literature is beyond our scope. In relation to problem decomposition 
\cite{kaelbling2010hierarchical} introduced Hierarchical Planning in the Now, which decomposes the problem strictly hierarchically, thereby limiting the need for long-term decision making, while still being suitable for long-horizons.
In \cite{garrett2015ffrob} a highly scalable symbolic planner is incorporated in the TAMP framework to achieve longer planning horizons.
However, currently proposed benchmarks lack very long-horizon problems~\cite{lagriffoul2018platform}, and to the best of our knowledge, there has not been any long-horizon application comparable to the one investigated in this paper demonstrated so far.

While simplifying problems is a common approach to make problems tractable in path planning~\cite{sharon2015conflict} and optimization~\cite{boyd2011distributed} -- the two areas from which we leverage methods -- it is not yet investigated in detail within TAMP settings.
In \cite{dogar2019multi} a team of multiple robots assembles a chair, where the planning problem is simplified by decomposing the problem into smaller subproblems by regrasping.
\cite{19-driess-RSSws} suggests the incorporation of hierarchies to solve a problem with less decisions before moving on to the actual problem, while noting that the worst-case scenario is still the same.

\subsection{Robotics in Fabrication, Architecture, and Construction}

As the interest in autonomous construction increases,~\cite{ardiny2015autonomous} identifies some challenges (e.g., cluttered working environments, high reliability to be useful, uncertainty in sensing) and current use cases (e.g., bricklaying, masonry).

Computational and robotic approaches in architecture and design are becoming more relevant as well and are often already deployed in reality:
\cite{adel2018design} discusses the assembly of specially designed wooden structures using two robotic arms for precise positioning.
The application of a mobile robot for semi-autonomous construction of a brick wall at the DFAB house, a demonstrator building for research concepts, is presented in \cite{Dorfler2016}.
In~\cite{willmann2012aerial}, flying vehicles are used for the construction of a tower consisting of foam-bricks in an art project.
In~\cite{bock2007construction}, use-cases in prefabrication and various mobile robots are presented, while noting that few of them are currently economically viable due to their specialized use cases.

While demonstrating use cases of robots in construction, these projects lack the explicit use of incorporating the robotic planning in the design process.
The robots are used for the construction, and made fit, rather than informing the design of what is feasible, which is what we are aiming for.



\section{BACKGROUND}

\subsection{Logic-geometric programming}

In this section, we formulate Logic-Geometric Programming (LGP) \cite{toussaint2017multi, toussaint2018differentiable}
as the underlying TAMP framework for the purpose of this work.
The main idea of LGP is to introduce a discrete variable $s\in\mathcal{S}$
(the state of a symbolic domain $\mathcal{S}$) that parameterizes the costs and
constraints of a nonlinear trajectory optimization problem (NLP) over the (continuous) path $x$.
Let $\mathcal{X} \subset \mathbb{R}^n\times SE(3)^{m}$ be the configuration space 
of $m$ rigid objects and $n$ articulated joints of potentially multiple agents
with initial condition $x_0$.
Given a goal description $g$ (details in \cref{sec:subgoals}), LGP tries to find a
sequence of symbolic states $s_{1:K}$, which we call a \emph{skeleton}, and a
path $x:[0,KT]\rightarrow\mathcal{X}$ in the configuration space that minimizes
\begin{align} \label{eq:LGP}
\!\!\min_{ x, K, a_{1:K}, s_{1:K} }   & \int_{0}^{KT}\!\!\!\!\!\!\!\!\! f_{\textrm{path}}(\overline{x}(t)) + \psi({x}(t), s_k) \diff t + f_{\textrm{goal}}(x(KT), g)  \notag\\	
\text{s.t.~}~~~~ & x(0)=x_0,~s_0 = \tilde{s}_0, ~h_\text{goal}(x(KT), g)  \notag\\
& \forall{t\in[0, KT]}: 
~~~h_{\textrm{path}}\big(\overline{x}(t), s_{k(t)}\big)=0  \notag\\
&~~~~~~~~~~~~~~~~~~~~~g_{\textrm{path}}(\overline{x}(t), s_{k(t)}) \leq 0 \notag\\
& \forall{k\in{1, \dots, K}}:~h_{\textrm{switch}}(\hat{x}(t), a_k)=0 \notag\\
&~~~~~~~~~~~~~~~~~~~~~g_{\textrm{switch}}(\hat{x}(t), a_k) \leq 0 \notag\\
&~~~~~~~~~~~~~~~~~~~~~a_k \in \mathcal{A}(s_{k-1})\notag\\
&~~~~~~~~~~~~~~~~~~~~~s_k \in \text{succ}(s_{k-1}, a_k)\notag\\
&~~~~~~~~~~~~~~~~~~~~~s_K \in \mathcal{G}(g), 
\end{align}
where $\overline{x} = (x, \dot{x}, \ddot{x})$ and $\hat{x} = (x, \dot{x})$.
This path consists of $K \in \mathbb{N}$ phases (kinematic modes) induced by the
sequence $s_{1:K}$, each of length $T>0$.
The number of phases is part of the optimization problem itself.
We assume the path to be globally continuous and two times continuously
differentiable within each phase.
$f_\text{path}$ describes path costs, e.g., squared accelerations.
The constraints $h_\text{path}$ and $g_\text{path}$ in phase $k$ of the path
are parameterized by $s_k$ with $k(t) = \lfloor t/T \rfloor$.
Therefore, the symbolic state determines the objective in each phase, while the
optimization problem tries to find a globally consistent path fulfilling these
requirements.
The transitions of $s_{k-1}$ to $s_k$ are determined by a first-order logic
language (similar to PDDL) through $\text{succ}(\cdot, \cdot)$ as a function of 
$s_{k-1}$ and the discrete action $a_k\in\mathcal{A}(s_{k-1})$.
The two functions $h,g_\text{switch}$ impose transition conditions between the 
modes.
$\tilde{s}_0\in\mathcal{S}$ is the initial symbolic state.
The goal is defined symbolically by $s_K \in \mathcal{G}(g)$ and kinematically
by $f_\text{goal}$ and $h_\text{goal}$ for the goal specification $g$ (see \cref{sec:subgoals} for more details).
We extend the formulation of \cite{toussaint2018differentiable} to include a
stability cost function $\psi$, which we will explain in \cref{sec:ranking}.


\subsection{Multi-Bound Tree Search}
The logic introduces a decision tree, where each leaf node, i.e., a node that
reaches a symbolic goal state $s_K\in\mathcal{G}(g)$, corresponds to an NLP as a
candidate for the overall TAMP problem.
Solving the LGP \eqref{eq:LGP} therefore includes a tree search over the discrete actions such that a symbolic goal state is reached and the NLP is feasible.
Using the full path optimization problem \eqref{eq:LGP} is too expensive to guide the tree search.
Therefore, a key contribution of \cite{toussaint2017multi} is to introduce relaxations $\mathcal{P}_i$ of \eqref{eq:LGP} in terms of lower bounds, i.e., computationally simpler problems that serve as a lower bound on the cost of the original problem and as a necessary condition on the feasibility.
For a given skeleton, \eqref{eq:LGP} is solved by discretizing $x$ in time.
This work relies on the following three bounds proposed in \cite{toussaint2017multi}:
The bound $\mathcal{P}_\text{pose}$ optimizes only the final pose for $t = KT$, and
$\mathcal{P}_\text{seq}$ the switching poses, i.e., $x$ is discretized with configurations corresponding to the switching times $t = 0, T, \ldots, KT$ only.
The full path optimization problem is called $\mathcal{P}_\text{full}$, where $x$ is discretized with $n_t$ points for each phase.

For scaling LGP to an architectural setting, we additionally introduce two new bounds: a stability bound and a sampling-based motion-planner bound in \cref{sec:ranking} and \cref{sec:mot_planning}, respectively.


\section{SCALING LGP TO ARCHITECTURAL CONSTRUCTION}

\subsection{Overview}

LGP is a very general way to formulate sequential manipulation problems. However, the existing solver (MBTS) is fundamentally limited in scalability: while it leverages approximate bounds to guide tree search over skeletons, it eventually always tries to optimize the full joint path over the whole sequence.
In our architectural construction application we need a long sequence of steps, i.e., in the order of thousands of actions $a_k$, to complete the full task -- optimizing jointly across that many steps is not scalable.
Further, while the optimization based approach provides many benefits (e.g., in dealing with higher dimensional multi-robot systems jointly, or optimizing interactions such as handovers jointly with the motions), the non-convexity of the motion problems ultimately lead to local optima.
To robustly solve large scale problems we need better guarantees of the sub-problem solvers: We cannot tolerate that a sub-problem is labelled infeasible only because an optimizer is trapped in a local optimum.
Sampling-based methods can help to gain robustness for such non-convex problems.

In this section we describe key aspects of our solver to address such challenges:
\begin{enumerate}
\item We propose an iterative limited-horizon approach to solve long-horizon LGPs, where the formulation of subgoals is a key aspect.
\item We propose a sampling-based motion-planner bound and stochastic restarting to address the challenges of non-convexity.
\item We propose a novel approach to approximate the static stability of the construction, leveraging a constrained optimization solver, which is used as an additional bound to guide tree search.
\end{enumerate}
The last point is specific to architectural construction, where the order of assembly should be guided by static stability.
To integrate this reasoning in our framework we need to have computationally efficient approximations, which we propose here.

\subsection{Subgoals \& Limited-Horizon LGP}\label{sec:subgoals}


We first propose to enable the planner to decompose the full problem into subgoals, each of which implies a limited-horizon LGP problem, akin to receding horizon planning.

Specifically, in our case, the overall goal specification $g = \bigcup_{i=1}^N \left\{(q_i, p_i)\right\}$ means that $N$ parts $q_i$ have to be placed at their specified target poses $p_i$, which translates to the constraint on the last symbolic state 
\begin{align}
s_K \in \mathcal{G}\left(\bigcup_{i=1}^N\left\{(q_i, p_i)\right\}\right),\label{eq:fullGoal}
\end{align}
where $\mathcal{G}(\cdot)$ extracts the symbolic goal state from the goal specification.
Instead of attempting to solve \eqref{eq:LGP} with the complete goal \eqref{eq:fullGoal}, we introduce a so-called horizon length $n_h$ and replace the symbolic goal constraint with
\begin{align}
s_K \in \bigcup_{ \left\{g_i \subseteq g ~:~ |g_i| = n_h\right\} }\mathcal{G}(g_i).\label{eq:subGoal}
\end{align}
While with the constraint \eqref{eq:fullGoal} a symbolic goal state is only reached if \emph{all} parts are placed, the subgoal formulation \eqref{eq:subGoal} terminates if at most $n_h$ parts are placed.
Therefore, the optimization horizon is automatically limited.
The overall goal is then achieved by solving \eqref{eq:LGP} with the constraint \eqref{eq:subGoal} iteratively for a specified horizon length $n_h$.
Parts that have already been placed in previous iterations are, of course, excluded from the goal specification in the current iteration.
Note that neither the subgoals nor their order is given explicitly, but subject to the planner itself.
This requires solving each subproblem robustly, while ensuring that those solutions result in physically and kinematically feasible intermediate goals.

\subsection{Static Stability Bound}\label{sec:ranking}
In our construction setup, we additionally have to prioritize which parts of the building should be placed first based on a stability criterion of the already placed parts.
This is expressed with the static stability term $\psi: (\mathcal{X}, \mathcal{S})\to\mathbb{R}$, which is added to the path costs in \eqref{eq:LGP} as a function of the current configuration $\mathcal{X}$ and the symbolic state $s$.
$\psi$ contributes to the costs only if in the symbolic state $s_k$ a part is placed.

While existing applications of LGP mainly focused on solving the underlying TAMP problem, i.e., finding a feasible solution, here we explicitly want to minimize this additional cost term.
In order to realize this efficiently, we introduce another lower bound, the stability bound $\mathcal{P}_\textrm{stab}$.
This bound evaluates $\psi$ for each subgoal, i.e., placement of a part, without taking into account any other constraints or costs that are induced by parts of the configuration space not corresponding to placed objects. 
Note that $\mathcal{P}_\textrm{stab}$ is only a lower bound on the cost and not a necessary condition for feasibility.

\subsection{Sampling-Based Motion Planning Bound}\label{sec:mot_planning}
Solving the full path optimization problem $\mathcal{P}_\textrm{full}$ in our setting is challenging due to collision avoidance in the complex geometries of our building scenario.
Therefore, we introduce another lower bound on $\mathcal{P}_\textrm{full}$ that utilizes sampling-based motion planning algorithms, specifically RRTs, to solve this issue.
This bound relies on the solution of the switching configurations from $\mathcal{P}_\textrm{seq}$, which allows us to combine the advantages of jointly optimizing these configurations including multi-agents and handovers with the path finding capabilities of the sampling-based planner.
In this work, we use RRT-connect \cite{kuffner2000rrt}
, and thus call the bound $\mathcal{P}_{\mathrm{RRT}}$.

Let $x^\textrm{seq}$ be the solution of the sequence bound $\mathcal{P}_\textrm{seq}$.
The bound $\mathcal{P}_\textrm{RRT}$ now aims to connect each consecutive switching configurations $x_{k-1}^\textrm{seq} = x^\textrm{seq}((k-1)T)$ and $x_{k}^\textrm{seq} = x^\textrm{seq}(kT)$ for $k=1, \ldots, K$ by solving
\begin{align} 
  \text{find}\ &x_\textrm{RRT}:[(k-1)T, kT] \rightarrow \mathcal{X} \notag\\
  \text{s.t.}\ &x_\textrm{RRT}((k-1)T) = x_{k-1}^\textrm{seq},\quad x_\textrm{RRT}(kT) = x_{k}^\textrm{seq} \notag\\
  &\forall t\in[(k-1)T, kT]: g_\text{path}(x_\textrm{RRT}(t))\leq 0 \notag\\
  &~~~~~~~~~~~~~~~~~~~~~~~~~~h_\text{path}(x_\textrm{RRT}(t))=0. \label{eq:mot_planner}
\end{align}
%
%
Equality constraints need special attention in sampling based planners \cite{Kingston2018}.
In our setup, the path constraints $h_\textrm{path}$ and $g_\textrm{path}$ only consider $x$, not $\overline{x}$, i.e., we relax the dynamic constraints for the RRT bound.
We note that sampling-based planners have previously been integrated in logic based task planners directly as the solver for the arising motion planning problem~\cite{hauser2009integrating}, but not explicitly as a bound of \cref{eq:LGP}.


\subsection{Stochastic Restarts}\label{sec:rob_opt}
Since even the lower bounds for \cref{eq:LGP} are still non-convex, and local optima the solver converges to might be infeasible, we introduce stochastic restarting of the underlying optimizer (KOMO)~\cite{14-toussaint-KOMO}, which is used to solve the NLPs, with randomized initial conditions for better reliability.

Specifically, we use restarting when solving the problems $\mathcal{P}_\textrm{pose}$ and $\mathcal{P}_\textrm{seq}$.
In both cases, we do not sample the variables in the full dimensional space $\mathcal{X}$, but identify the most relevant variables (i.e., the ones that influence the point to which a solution converges to the most) in the NLP, randomize those, and project them up into the full configuration space (see \cref{sec:stoch_rest_imp_det} for an example).
The other variables are initialized from the current state as solved in the previous iteration.

One has to choose carefully how many restarts are allowed, before a given NLP is declared \textit{infeasible}.
We tackle this issue by introducing a computational budget $t_\textrm{max}$ for the solution-attempts, and declare a problem as infeasible in case no solution was obtained.

\subsection{Algorithm and Hierarchy of Bounds}
This section describes how the bounds are used to efficiently solve the LGP problem for each subgoal and hence the overall long-term TAMP problem.

By construction, the hierarchy of our bounds is
\begin{align}
    \mathcal{P}_\textrm{stab} \prec \mathcal{P}_\textrm{pose} \prec \mathcal{P}_\textrm{seq} \prec \mathcal{P}_\textrm{RRT} \prec \mathcal{P}_\textrm{full},\label{eq:boundHierarchy}
\end{align}
where $\prec$ means a lower bound both in terms of costs and as a necessary condition of the feasibility of a higher bound.
Let $\mathcal{B}$ be the set of all symbolic state sequences that reach the symbolic goal constraint \eqref{eq:subGoal} for horizon length $n_h$.
The overall idea of the algorithm is to select $s_{1:K}\in\mathcal{B}$ and then test its feasibility by computing the bounds \eqref{eq:boundHierarchy} until either one bound is infeasible or $\mathcal{P}_\text{full}$ is feasible and hence a solution has been found.

However, due to the combinatorial complexity, determining $\mathcal{B}$ fully may not be possible even for short horizon lengths.
To obtain possible action sequences, the logic-tree is expanded by breadth-first search until a sequence $s_{1:K}$ is reached that fulfills \eqref{eq:subGoal}, and $\mathcal{P}_\textrm{stab}$ is satisfied as detailed in \cref{sec:stabilityOptimization}.
For the best candidate according to $\mathcal{P}_\mathrm{stab}$ we then attempt to compute the pose and sequence bounds, $\mathcal{P}_\textrm{pose}$ and $\mathcal{P}_\textrm{seq}$, respectively.

According to \eqref{eq:boundHierarchy}, the RRT-bound $\mathcal{P}_\textrm{RRT}$ is a lower bound to $\mathcal{P}_\textrm{full}$, and should thus be computed first.
However, since the worst-case run-time of $\mathcal{P}_\textrm{RRT}$ is only limited by a computational budget, which has to be high enough in order not to miss feasible solutions, solving $\mathcal{P}_\textrm{RRT}$ can get expensive.
In comparison, $\mathcal{P}_\textrm{full}$ either gives a solution or returns infeasibility much faster.
Since a local optimizer is not guaranteed to find a solution for a path between the poses obtained from $\mathcal{P}_\textrm{seq}$ due to non-convexities, the `infeasibility' assignment by $\mathcal{P}_\textrm{full}$ contains false positives.
Therefore, we first solve $\mathcal{P}_\textrm{full}$ (initialized with $\mathcal{P}_\textrm{seq}$) and only if this returns infeasibility we utilize the RRT bound, which, in case it can find a solution, is used as an initialization for a final smoothing step.


\subsubsection{Details about stability optimization}\label{sec:stabilityOptimization}
 %
%
%
Ideally, one would solve
\[\arg\!\min_{s_{1:K}}\ \sum_{k=1}^{K}\psi(x(kT), s_k) \ \textrm{s.t.}\ s_{1:K}\in \mathcal{B}\]
to decide which sequence $s_{1:K}$ to evaluate further. 
However, we only have access to the set $\mathcal{B}'\subseteq \mathcal{B}$, which is grown through the expansion of the tree.
To balance the tree expansion and the optimality of the solution, we propose a scheme that provides such a trade-off, by proceeding with the solution of $\mathcal{P}_\textrm{pose}$ if
\begin{align}\sum_{i=1}^{K_1}\psi(x^{(1)}(iT), s_i^{(1)})-\sum_{j=1}^{K_2}\psi(x^{(2)}(jT), s_j^{(2)})<\epsilon(|\mathcal{B}'|), \notag\end{align}
where $s^{(1)}_{1:K_1}$ and $s^{(2)}_{1:K_2}$ are the best and the second best sequence found so far that reach the symbolic goal \eqref{eq:subGoal}, respectively.
$\epsilon$ is a function of the number of currently found sequences. 
We additionally impose a minimum size of the set $\mathcal{B}'$, and a maximal computational budget for the expansion.
The choice of $\epsilon$ is dependent on the behavior of $\psi$, and the tolerable sub-optimality for the specific problem.



\section{EXPERIMENTS \& RESULTS}


\subsection{The BUGA Wood Pavillon}

The BUGA Wood pavilion is made of 376 unique wooden elements which are precisely fabricated by robots.
It spans 30 meters, and was assembled by two human operators and one crane in 10 days.
For the fabrication of the timber-parts, robotic constraints were taken into account, whereas the construction was manually planned.

In this work, we scale the elements of the pavilion to 80\% of their true size to avoid difficulties for the motion planning algorithm when placing the part at its final position, as this positioning is not the main focus of this paper.

\subsection{Robots}
\begin{figure}
    \centering
    \includegraphics[width=0.3\textwidth]{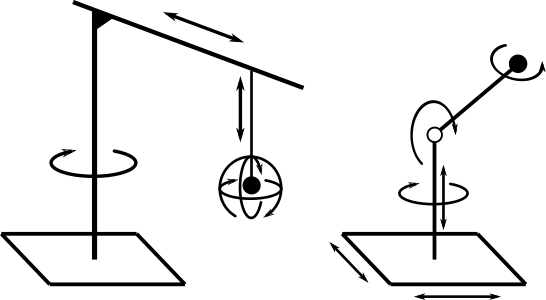}
    \caption{Schematics of the crane with 6 DoF, and the mobile robot with 6 DoF used in the problem setting.}
    \label{fig:robots}
\end{figure}

In this problem setting, we use two robots with different sets of capabilities (\cref{fig:robots}), which mimic how the construction process was done in reality before:
\begin{itemize}
\item One crane, which is used to lift the parts, and move them to the handover position.
The rotation of the end-effector of the crane around the x and y axis is limited to only allow for small rotations.

\item One robot with a mobile base, used for final positioning and placement of the parts.
The translation on the z-axis is limited in height, and the rotation of the arm is limited to not bend further down than the joint position.
\end{itemize}



\subsection{Assumptions}
We assume that the final pose and position of each part, and thus the whole design of the structure is known before the construction process starts.
Hence, the termination criterion of the planner is that each piece has to be placed at its corresponding final position.
We also assume:

\begin{itemize}
    \item No noise is present in the movement for both the crane and the mobile robot,
    \item Stable grasps by touch for both the crane and the mobile robot.
\end{itemize}

\subsection{Modeling details}
\subsubsection{Logic}
We describe the details of the used logic predicates for the specific task of building the BUGA wood pavilion.
Most of the implementation details are the same as in \cite{toussaint2017multi, toussaint2018differentiable}, with small changes to account for the differences in robots and task domain:
The predicates, their implicitly imposed path constraints and the action operators are listed in \Cref{Tab:Predicates} and \Cref{Tab:Actions}, respectively.
They are slightly different from the ones used in a normal pick-and-place problem, as we impose that the part has to be placed from above, such that the mobile robot is only `finalizing' the position of the part.

\subsubsection{Subgoals}
We define the subgoals $g_i$ as having placed $M$ ($M\ll N$) parts $q_j$ of the structure successfully, which translates to a horizon length of $n_h = M$.
This translates to a minimum of $4\times M$ necessary actions to fulfill the subgoal.
The final goal $g$ is the full assembly of the whole building, i.e., all parts being placed by the robots.

We introduce an adaptive approach of choosing the limited horizon length: We start the algorithm with $n_h=n_\text{max}$, and decrease it if within a computational budget no feasible solution has been found. 
In case we were able to solve the problem for two consecutive subgoals with horizon length $n_h$, we increase the horizon length given that $n_h<n_\textrm{max}$.

\begin{table}[t]
\centering
\caption{Predicates to impose constraints on the path optimization}
\begin{tabular}{>{\hspace{0pt}}p{0.312\linewidth}|>{\hspace{0pt}}p{0.684\linewidth}}
(touch X Y)         & distance between X and Y equals 0                                                                                   \\
(stable X Y)        & \begin{tabular}[c]{@{}l@{}}create stable (constrained to zero velocity) free\\(7D) joint from X to Y\end{tabular}  \\
(postLift X Y)      & X is above Y with a distance \textgreater{} 0                                                                      \\
(preHandover X Y Z) & \begin{tabular}[c]{@{}l@{}}X is above Y, which is the goal of Z, with a \\distance \textgreater{} 0~\end{tabular}  \\
(fitPose X Y)       & pose of X (7D) is exactly at pose of Y                                                                            
\end{tabular} \label{Tab:Predicates}
\end{table}

\begin{table}[t]
\centering
\caption{Action operators and the path constraints they imply.}
\begin{tabular}{l|l}
(pick X Y)       & {[}touch X Y] (stable X Y)                      \\
(liftUp X Y)     & (postLift X Y)~                                 \\
(handover X Y Z) & {[}touch X Y] (stable X Y) (preHandover X Y Z)  \\
(place X Y)      & {[}stable Y X] (fitPose X Y)                   
\end{tabular} \label{Tab:Actions}
\end{table}

\begin{figure*}
\centering
\begin{subfigure}[t]{.31\textwidth}
  \centering
  \includegraphics[width=\linewidth]{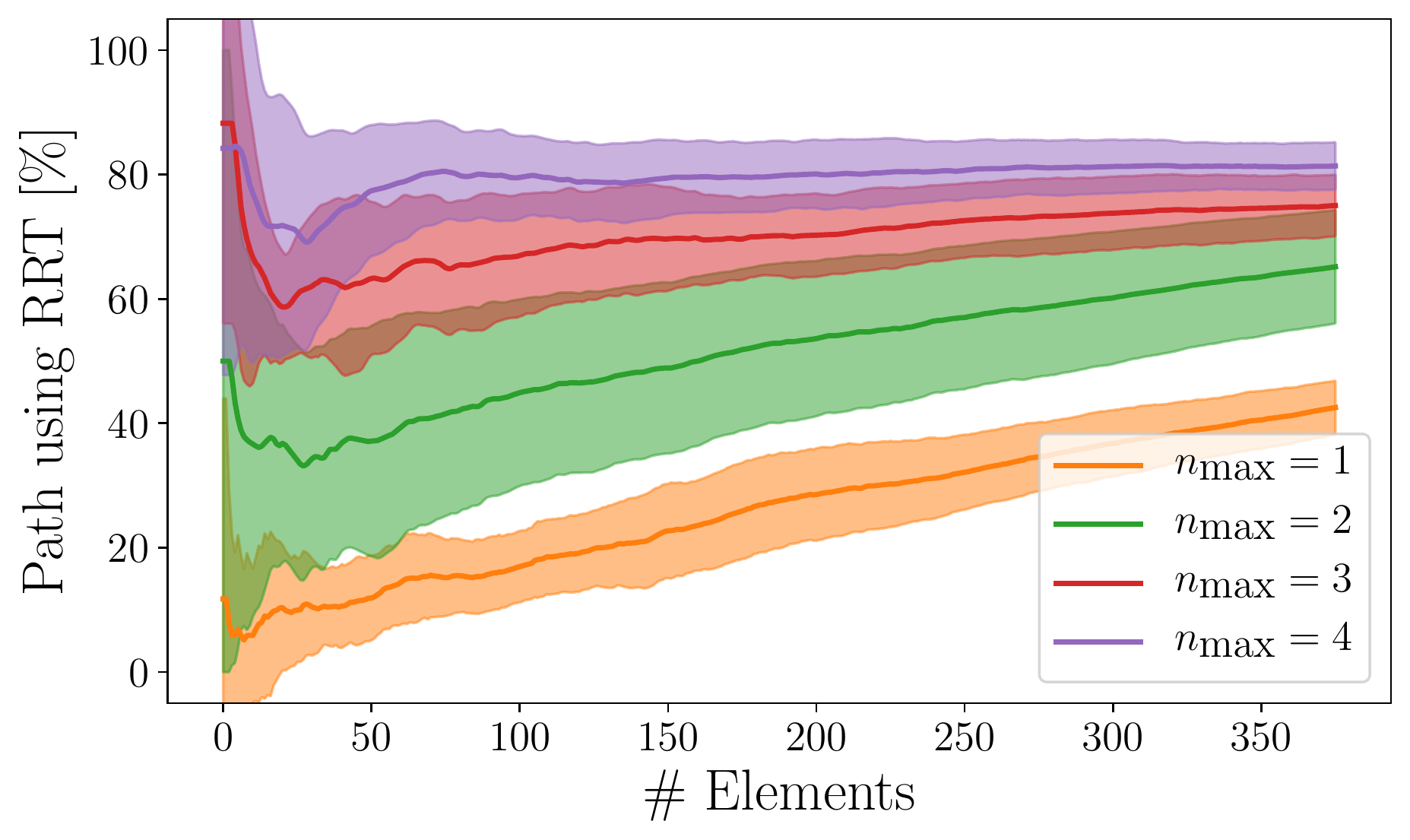}
  \caption{\label{fig:rrt_usage}Relative usage of the sampling-based planner throughout runs for different $n_{\textrm{max}}$.}
\end{subfigure}%
\hspace{1em}
\begin{subfigure}[t]{.31\textwidth}
  \centering
  \includegraphics[width=\linewidth]{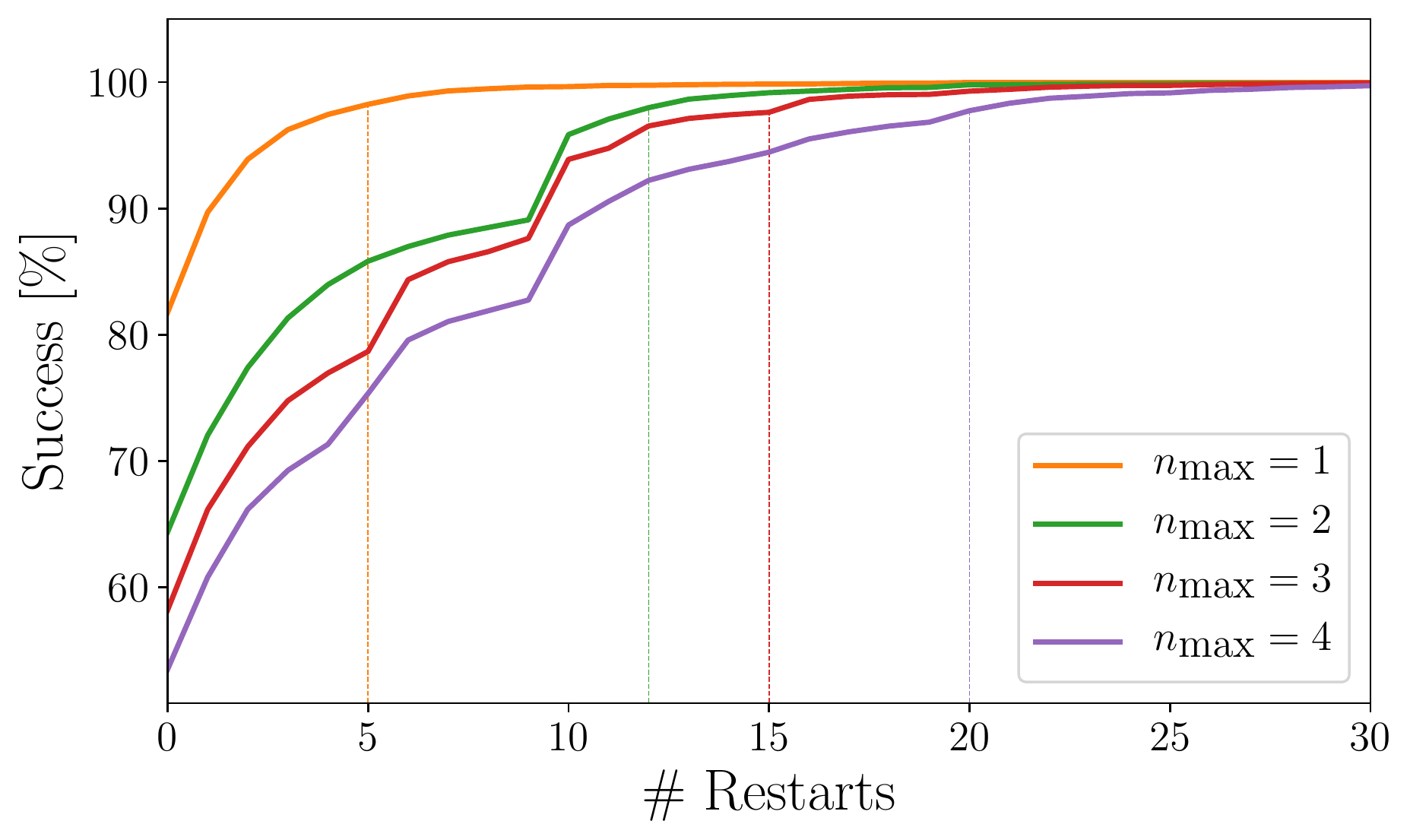}
  \caption{\label{fig:restarts}Average number of required restarts before finding a solution to $\mathcal{P}_\textrm{Seq}$. The dashed line shows where at least a success rate of 97.5\% is attained.}
\end{subfigure}
\hspace{1em}
\begin{subfigure}[t]{.31\textwidth}
    \centering
    \includegraphics[width=\textwidth]{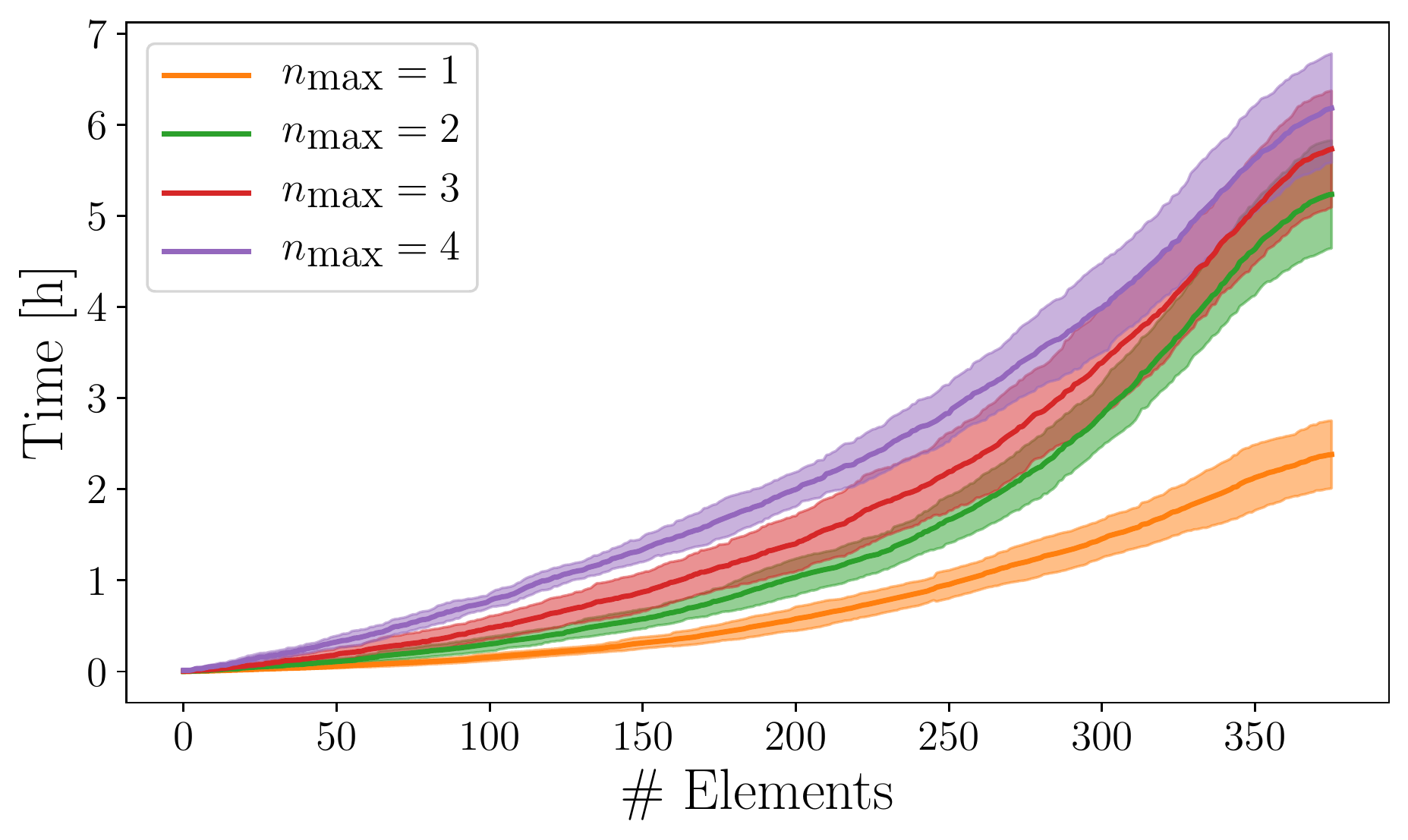}
    \caption{\label{fig:time}Total computation times for different horizon-lengths $n_h$.}
\end{subfigure}
\caption{Usage of the sampling-based planner, required restarts, and computation time for $\psi_{\textrm{neighbor}}$ and different $n_h$.}
\label{fig:robustness}
\end{figure*}

\subsubsection{Stability bound}
We define two stability functions $\psi_\text{neighbors}(x,s)$ and $\psi_\text{statics}(x,s)$. From $x$ we determine the set $\mathcal{U}$ of previously placed parts, and from $s$ the set $\mathcal{W}$ of potentially placeable parts in the evaluated action sequence.
We additionally define $\mathcal{N}(w)$ as the set of neighbors of part $w$:
\begin{itemize}
	\itemsep0em
	\item $\psi_\text{neighbors} = -\frac{1}{|\mathcal{W}|}\sum_{w\in |\mathcal{W}|}|\mathcal{N}(w)\cap \mathcal{U}|$, i.e., the average number of neighbors per part.
	\item $\psi_\text{statics} = \frac{1}{|\mathcal{U}\cup\mathcal{W}|} \mathtt{Torques}(\mathcal{U}\cup\mathcal{W})$, i.e., we compute the sum of all torques between neighboring parts, and take the average over the building to normalize for number of placed parts.
\end{itemize}

\subsubsection{Stochastic restarts}\label{sec:stoch_rest_imp_det}
For the mobile robot, we randomly sample the translational coordinates, and do not change the other variables.
Changing these coordinates leads to converging to the solutions of the switching-positions from different sides of the pavilion, which helps the optimizer avoid getting stuck in an infeasible configuration.
For the crane, no variables are randomized.


\subsection{Experimental results}
In the following section, we analyze several different scenarios, and provide detailed analysis to demonstrate the necessity and the effectiveness of our novel extensions on the default LGP formulation.
If not stated differently, the experiments\footnote{The experiments were run on a single core of Intel(R) Xeon(R) Gold 6148 CPU @ 2.40GHz per experiment.} were run several times with different random seeds and the evaluated metrics were averaged to reduce the stochastic effects of the sampling and restarting.
A video of a full run with $n_h=1$ and $\psi_\text{neighbors}$ is part of the supplementary material.

\subsubsection{Robustness}
We report the percentage of times the sampling-based planner was invoked, and the required restarts over the whole run in \cref{fig:rrt_usage}, and \cref{fig:restarts}, respectively. 
As the environment gets harder for the optimizer to solve for a motion path (i.e., a solution to $\mathcal{P}_\mathrm{full}$ using KOMO is not found), especially with higher horizon lengths, our framework reliably switches to the RRT-based motion planner. 

\subsubsection{Stability function}
\begin{figure*}
\centering
\begin{subfigure}[t]{.2\textwidth}
  \centering
  \includegraphics[width=.93\linewidth,angle=73]{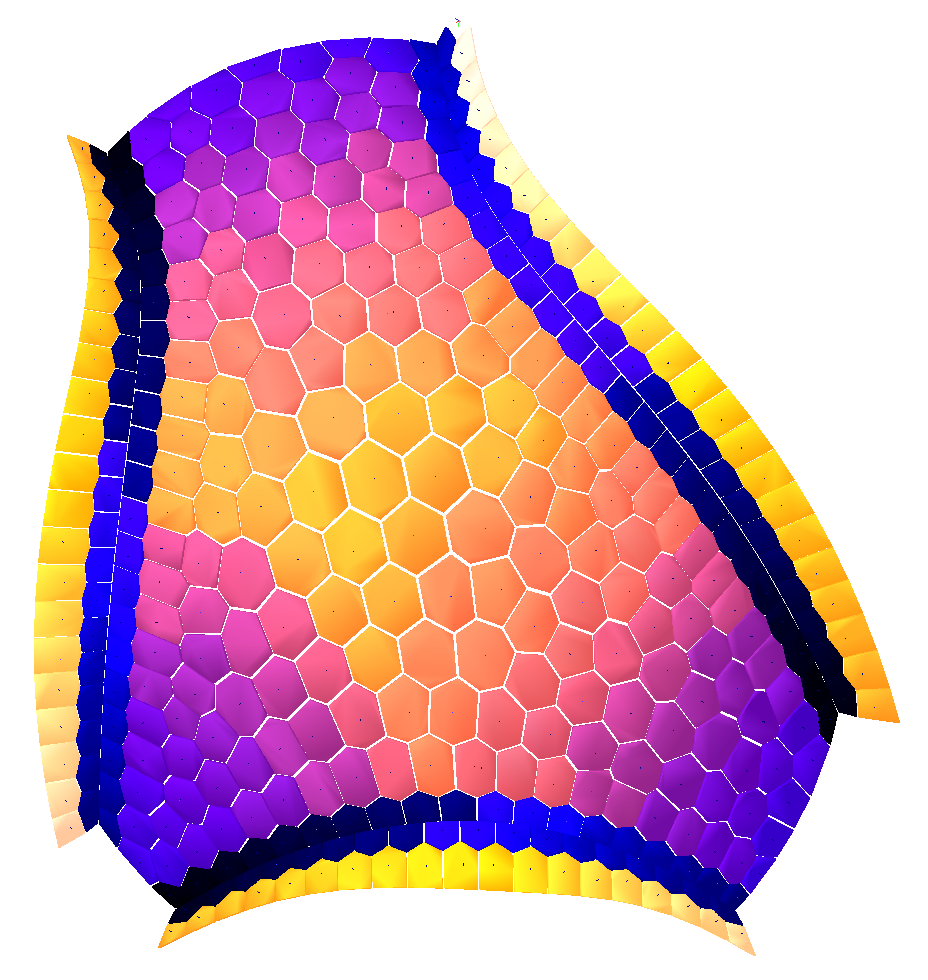}
  \caption{\label{fig:seq_0}The assembly order used in reality.}
\end{subfigure}
\hspace{1em}
\begin{subfigure}[t]{.2\textwidth}
  \centering
  \includegraphics[width=.93\linewidth,angle=73]{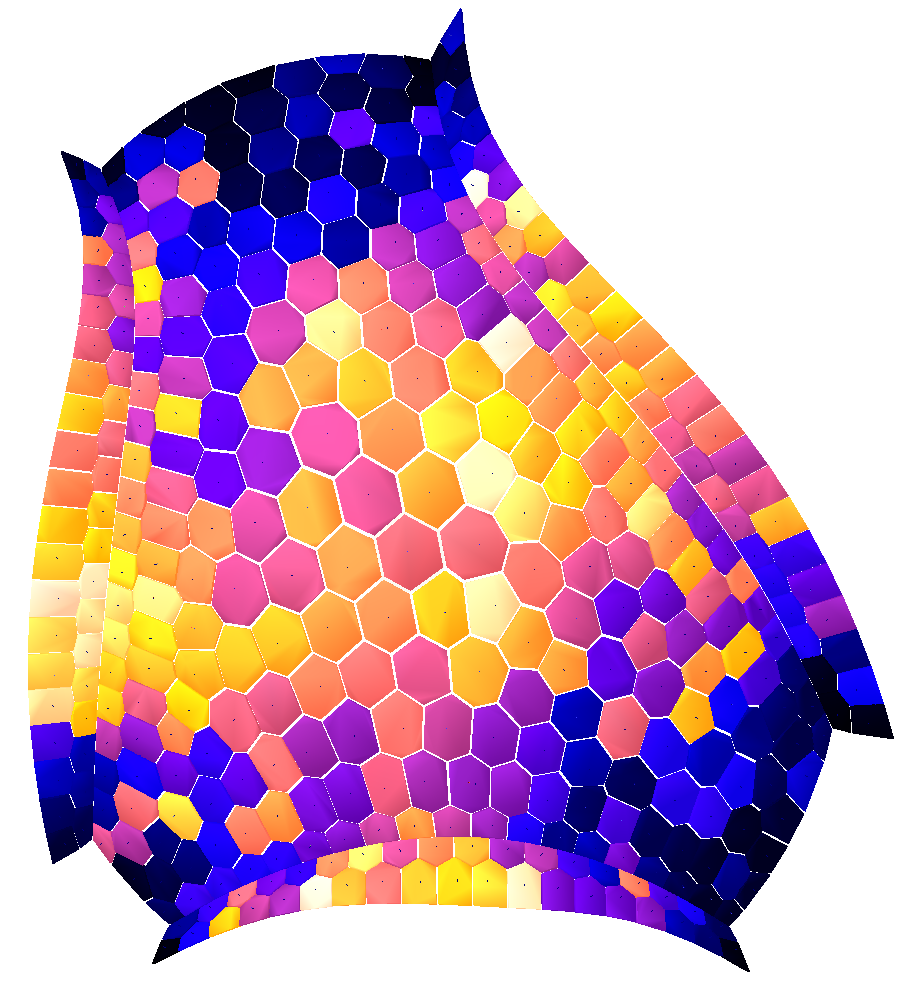}
  \caption{\label{fig:seq_1}Without optimization function.}
\end{subfigure}
\hspace{1em}
\begin{subfigure}[t]{.2\textwidth}
  \centering
  \includegraphics[width=.95\linewidth,angle=73]{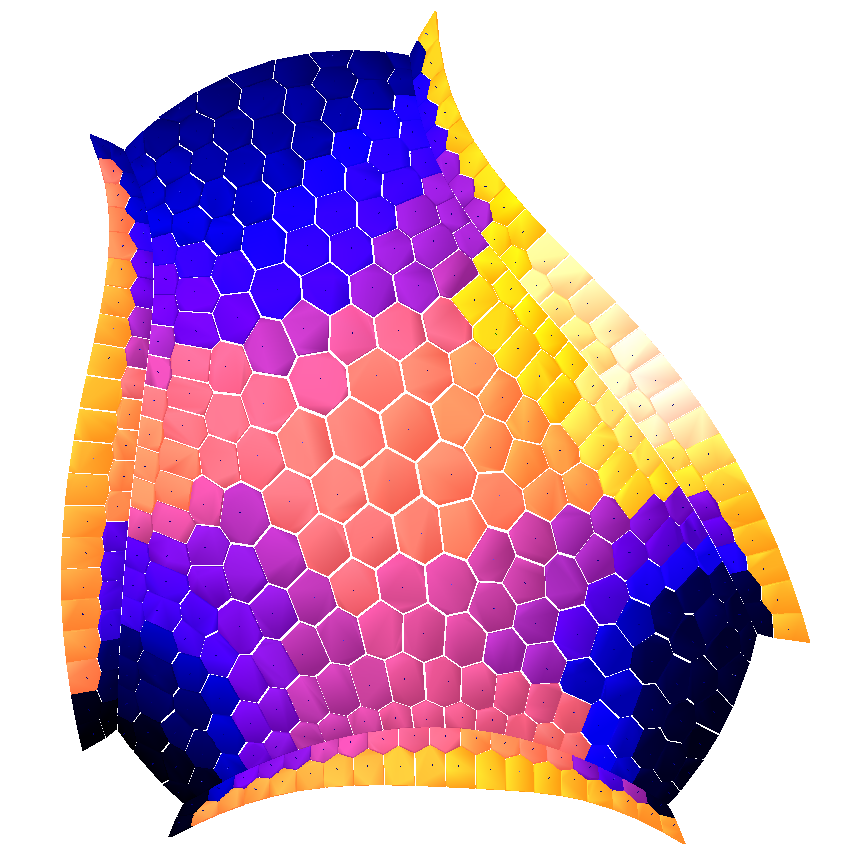}
  \caption{\label{fig:seq_2}Using $\psi_{\text{neighbor}}$.}
\end{subfigure}
\hspace{1em}
\begin{subfigure}[t]{.2\textwidth}
  \centering
  \includegraphics[width=.93\linewidth,angle=73]{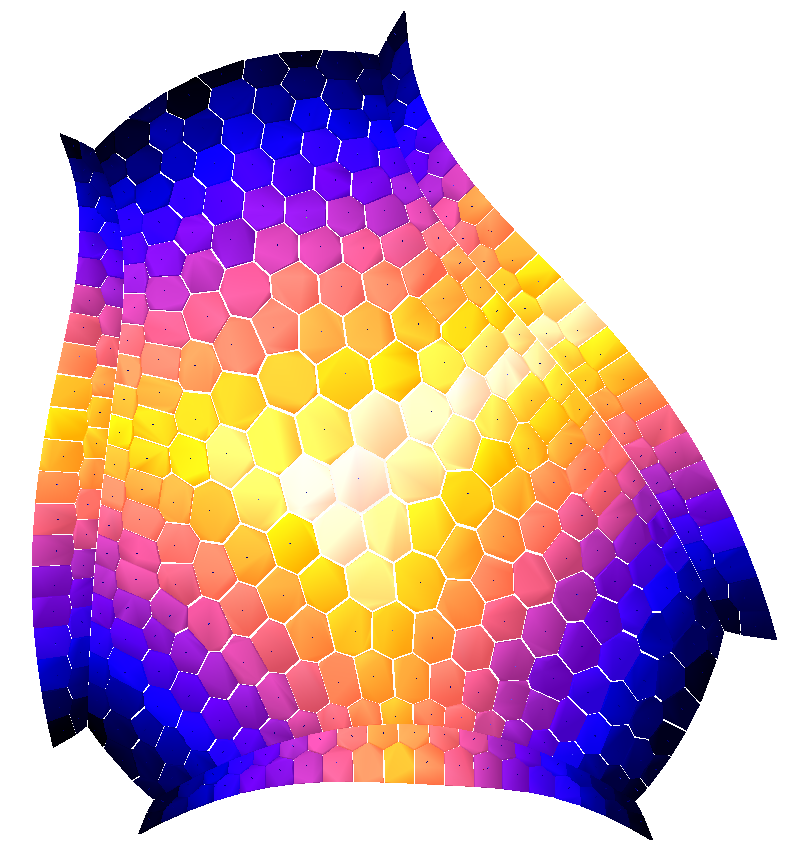}
  \caption{\label{fig:seq_3}Using $\psi_{\text{statics}}$.}
\end{subfigure}
\caption{Sequences arising from the different stability criteria (dark parts are placed first, light parts at the end) with horizon length $n_h = 1$.}
\label{fig:sequences}
\end{figure*}

The results of applying different strategies highlight the capability of the developed framework to enable an exploratory functionality that can support users, and specifically architects for this use-case.
In effect, different sequences arise with two different stability functions (\cref{fig:seq_2}, \cref{fig:seq_3}), compared to the baseline without ordering (\cref{fig:seq_1}), where we only impose the existence of one connection to a neighbouring part as the sufficient condition.
The order used when assembled in reality is shown in \cref{fig:seq_0}.

\subsubsection{Horizon length}
We compare the required computation time between different horizon lengths $n_h$ over the course of a run (\cref{fig:time}).
We note that, while longer horizons ($n_h > 4$, which translates to a minimum of 20 logical actions) are theoretically possible, our analysis on this is limited due to computational constraints, namely restrictions in the logical search which exhibits combinatorial complexity.

\Cref{tab:summary} gives a summary of experiments over a range of horizon lengths $n_\textrm{max}$. In the following, we will discuss some of the results in more detail:

\begin{table}[]
\begin{center}
    \caption{Average time usage with different $\psi$ and $n_\mathrm{max}$.}
    \begin{tabular}{l@{\hskip 0mm}r|@{\hskip 1mm}
            S[table-format=3.2]@{\hskip 0mm}S[table-format=3.2]@{\hskip 0mm}|@{\hskip 1mm}
            S[table-format=3.2]@{\hskip 1mm}S[table-format=3.2]@{\hskip 1mm}S[table-format=3.2]@{\hskip 1mm}S[table-format=3.2]@{\hskip 1mm}|@{\hskip 1mm}
            S[table-format=3.2]@{\hskip 0mm}}
    & & & & \multicolumn{5}{c}{Time [h]} \\
    & & 
    {\multirow[b]{2}{*}{\shortstack{Restarts\footnotemark{}\\ \relax[\%]}}} & {\multirow[b]{2}{*}{\shortstack{RRT\\ \relax[\%]}}} & 
    {\multirow{2}{*}{\shortstack{Logic \&\\ $\mathcal{P}_\mathrm{Stab}$}}} & & &
    {\multirow{2}{*}{\shortstack{$\mathcal{P}_\mathrm{RRT}$ \&\\ $\mathcal{P}_\mathrm{Full}$}}} &  \\
    $\psi$ & $n_\text{max}$ & & & & {$\mathcal{P}_\mathrm{Pose}$} & {{$\mathcal{P}_\mathrm{Seq}$}} & & {Total}\\
    \hline
    \multirow{3}{*}{statics} 
                & 1 & 10.0 & 33.4 & 3.81 & 0.09 & 0.79 & 0.83 & 5.5 \\
                & 2 & 22.3 & 64.3 & 2.87 & 0.07 & 2.54 & 1.40 & 6.9 \\
                & 3 & 25.9 & 77.4 & 2.70 & 0.07 & 3.88 & 1.53 & 8.2 \\
    \hline
    \multirow{4}{*}{\shortstack{neigh-\\bors}} 
                & 1 & 17.9 & 42.3 & 0.02 & 0.12 & 1.03 & 1.21 & 2.4 \\
                & 2 & 28.7 & 65.1 & 0.01 & 0.09 & 3.42 & 1.71 & 5.2\\
                & 3 & 29.2 & 75.1 & 0.01 & 0.07 & 3.95 & 1.70 & 5.7 \\
                & 4 & 30.7 & 80.8 & 0.27 & 0.07 & 4.22 & 1.63 & 6.2 \\
    \end{tabular}
    \label{tab:summary}
\end{center}
\end{table}
\footnotetext{Percentage of optimization instances where a restart was necessary to obtain a feasible solution in the optimization problem.}

\begin{figure}
    \centering
    \includegraphics[width=0.4\textwidth]{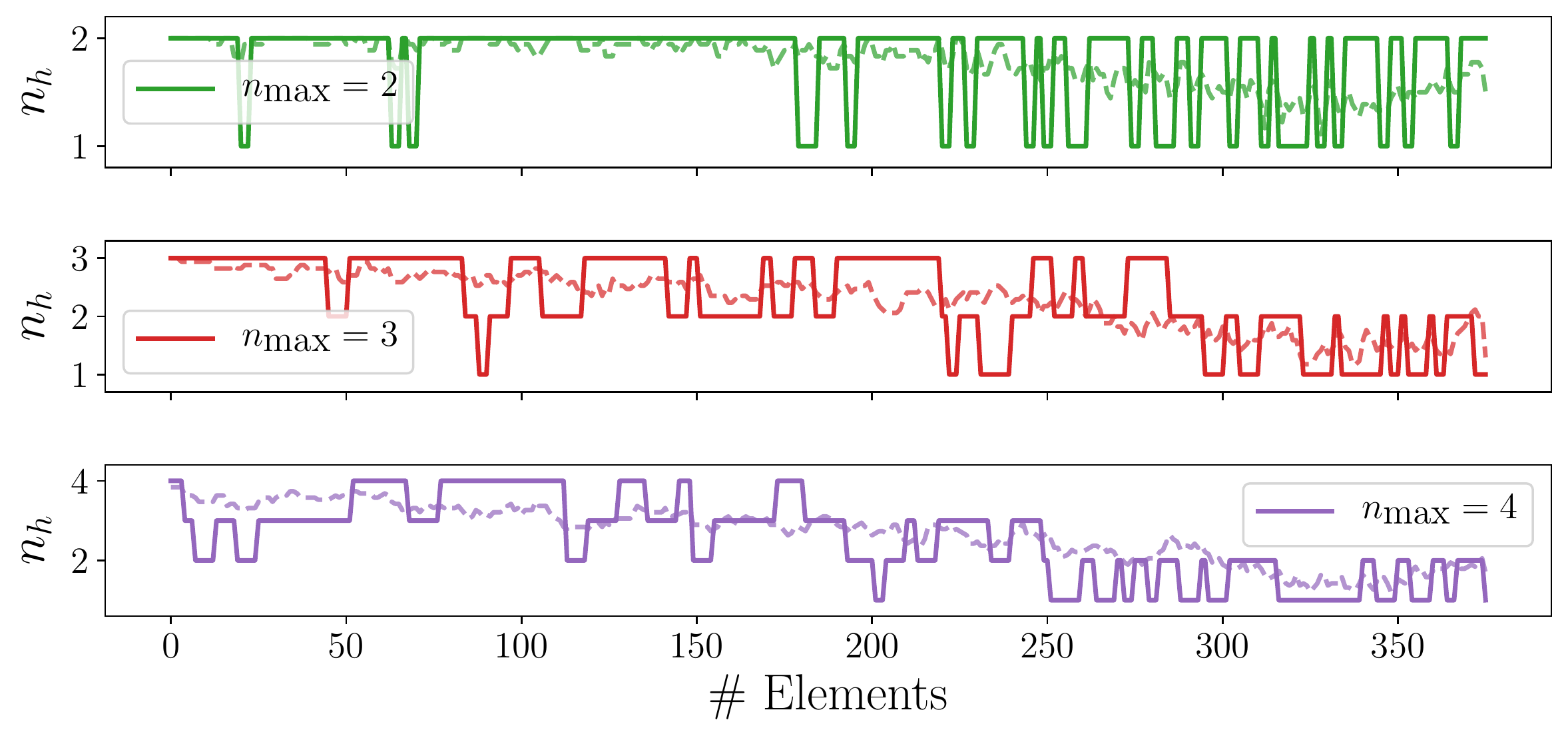}
    \caption{Average (dashed) and exemplary (full) trajectories of horizon length $n_h$ for various $n_{\text{max}}$ through a full run.}
    \label{fig:h_traj}
\end{figure}

\begin{itemize}
    \item Comparing RRT usage over horizon-length: The results indicate that the necessity of using a robust motion planner becomes more important when dealing with longer horizon lengths.
    This is intuitive, since optimization can not solve the whole path planning problem even if only one of the `sub-problems' is infeasible.
    Hence, the increased reliance on the RRT-planner from $n_h=1$ to $n_h=4$ is expected.
    \item Restarting compared over horizon length: While roughly 80\% of all problems for $n_h=1$ can be solved without any restarts, the utility of the restarting scheme becomes critical when dealing with larger $n_h$.
    \item Comparison of the stability functions: The stability function imposes the order of the parts (\cref{fig:sequences}). This is helpful for the feasibility of the following bounds in the MBTS, which in turn leads to a decrease in the necessary time for the \textit{sequence} ($\mathcal{P}_\mathrm{Seq}$) and \textit{motion} ($\mathcal{P}_\mathrm{RRT}$ \& $\mathcal{P}_\mathrm{Full}$) bounds.
    This comes with a stark trade-off of necessary computation time.
    In sum, the total computation times when using $\psi_{\text{statics}}$ are approximately 1.5 to 2 times longer.
    \item The switching scheme for the adaptive horizon is aggressive in trying to get back to the maximum allowed horizon length.
    It can be seen in \cref{fig:h_traj} that a longer horizon fails, and oscillatory behavior emerges between the short and the longer horizon length.
    This can potentially lead to wasted effort, which could be avoided by using a more suitable switching scheme.
\end{itemize}


\section{DISCUSSION}
\subsection{Limitations} \label{sec:limitations}
The limitations that arise in our demonstration can be grouped in two major areas: \textit{i}) issues that prevent scaling of this approach to other, larger problem instances:
\begin{itemize}
    \item Decomposing the long-horizon problem into a problem with multiple disjoint subgoals is not feasible for some problems that are more reliant on previous decisions of the TAMP problem.
    Similar to Model Predictive Control (MPC), the limited horizon approach we introduce here decreases the set of feasible solutions for future decisions.
    The stability-bound helps mitigate, but does not resolve possible challenges from this completely.
    \item Sampling-based approaches do not scale favourably over multiple agents in the naive compound-configuration-space formulation.
    Planning for multiple agents is still feasible and performant enough for the case of two agents, but scaling this approach to parallel construction using multiple agents on the same building is not feasible.
\end{itemize}
and \textit{ii}) caveats to the solutions that we introduced for the increase in robustness of the LGP approach:
\begin{itemize}
    \item Using sampling-based motion planners offers a solution to the problem that optimizers fail in cluttered environments, but they come with their own set of problems, namely not being able to declare a problem as infeasible, and computing paths that might be feasible when ignoring system-uncertainties, but are actually infeasible.
    \item The introduced method of restarting the optimizer currently still suffers from a similar issue as the sampling-based planner: in a non-linear optimization problem it is, in general, not known if a solution can be found, or if a different action sequence should be followed.
\end{itemize}


\subsection{Outlook}
Solving the issues stated in~\Cref{sec:limitations} would allow our approach to be scaled more efficiently to multiple agents, leading to more versatile construction processes.
A more intelligent scheme for the determination of the subgoals should be an area of further research.
The incorporation of future costs in the TAMP framework similar to the terminal set and the terminal cost in MPC might alleviate issues that could arise when choosing the subgoals for the decomposition improperly.

Neglecting the noise and dynamics arising in the real world makes many of the planned trajectories suboptimal or even infeasible in the real world.
Incorporating measures akin to the stability function for the controllability would enable more realistic planning.
Future work also targets incorporating a more structured sampling of the starting points, and a (better) stopping criterion, which have previously been investigated in optimization literature.

\section{Conclusion}
We introduced a novel TAMP solver incorporating a limited horizon length and sampling-based planning methods in combination with novel stability analysis to solve complex, long-horizon TAMP problems in the construction domain robustly.
Our work can be seen as a first step towards unifying assembly and construction planning, which in turn can enable co-design, i.e., the design is iteratively analyzed and improved in terms of geometric, kinematic, and static feasibility of the whole construction process.

We combine optimization and sampling-based methods to combine the strengths of both approaches.
We show that restarts of the optimizer makes our framework more robust, and thus applicable to complex, long-horizon TAMP problems.

\section*{Acknowledgements}
We thank Hans-Jakob Wagner and Long Nguyen for their help with the creation of the video, and for providing the model of the BUGA pavilion.


\bibliographystyle{IEEEtran}
\bibliography{references}

\end{document}